\documentclass[letterpaper, 10 pt, conference]{ieeeconf}  

\IEEEoverridecommandlockouts                              

\usepackage{amsmath}
\usepackage{amssymb}
\usepackage{lipsum}
\allowdisplaybreaks
\usepackage{amsmath}
\usepackage{graphics} 
\usepackage{graphicx}
\usepackage{bm}
\usepackage{subfigure}
\usepackage{eqnarraycompact}
\usepackage{mathtools}
\newcommand{\define}{\stackrel{\Delta}{=}}
\newcommand\bbR{\mathbb{R}}

\title{\LARGE \bf
DS-VIO: Robust and Efficient Stereo Visual Inertial Odometry based on Dual Stage EKF
}

\author{Xiaogang Xiong$^{1}$, Wenqing Chen$^{1}$, Zhichao Liu$^{2}$ and Qiang Shen$^{3}$
\thanks{$^{1}$ X. Xiong and W. Chen are with Faculty of Mechanical Engineering and Automation,
        Harbin Institute of Technology, Shenzhen, P.R. China
        {\tt\small xiongxg@hit.edu.cn}}%
 \thanks{$^{2}$ UBTECH Robotics, Shenzhen, China}       
\thanks{$^{3}$ Arizona State University, School for Engineering of Matter, Transport and Energy, USA
        {\tt\small qiang.shen@asu.edu}}%
}

\begin{document}

\maketitle
\thispagestyle{empty}
\pagestyle{empty}

\begin{abstract}

This paper presents a dual stage EKF (Extended Kalman Filter)-based algorithm for the real-time and robust stereo VIO (visual inertial odometry). The first stage of this EKF-based algorithm performs the fusion of accelerometer and gyroscope while the second performs the fusion of stereo camera and IMU. Due to the sufficient complementary characteristics between accelerometer and gyroscope as well as stereo camera and IMU, the dual stage EKF-based algorithm can achieve a high precision of odometry estimations. At the same time, because of the low dimension of state vector in this algorithm, its computational efficiency is comparable to previous filter-based approaches. We call our approach DS-VIO (dual stage EKF-based stereo visual inertial odometry) and evaluate our DS-VIO algorithm by comparing it with the state-of-art approaches including OKVIS, ROVIO, VINS-MONO and S-MSCKF on the EuRoC dataset. Results show that our algorithm can achieve comparable or even better performances in terms of the RMS error.

\end{abstract}

\section{INTRODUCTION}

In GPS-denied environments, for instance, indoors and in urban canyons, it is essential for mobile robot platforms such as micro aerial vehicles (MAVs) to know their own pose for operations. In recent years, this pose estimation problem is popularly solved by the combination of visual information from cameras and measurements from an Inertial Measurement Unit (IMU), which is usually referred to as Visual Inertial Odometry (VIO)\cite{Sun_2018_Robust,zheng2018pi}. Compared to lidar based approaches, VIO requires only a lightweight and small-size sensor package, making it the preferred technique for mobile robot platforms.


 Realtime VIO solutions are critical for robots to response quickly to corresponding environments. At the same time, computation efficiency of VIO solutions is also important for robots with limited payloads and computation power such as MAVs. It is well-known that filter-based VIO solutions are generally more computationally efficient than that of optimization based methods \cite{Sun_2018_Robust} while stereo cameras are more robust to hostile environments compared with monocular camera. In order to solve the above problem of pose estimation in a realtime and robust manner, we propose a new VIO solution based on the extended Kalman filer (EKF) and information from stereo cameras in this paper.

\par

To achieve a precise and robust pose estimation, the algorithm we present here is a dual stage EKF-based VIO solution, which is referred as to DS-VIO. It is consisted of two EKF filters and each stage has one EKF filter. We demonstrate that our algorithm perform comparable or even higher precision than other state-of-art VIO algorithms. The primary contribution of our work is a new filter-based VIO framework that contains a dual stage EKF filter. The first stage perform the sensor fusion of the measurements of accelerometer and gyroscope. The second stage performs the fusion of the measurements of IMU and stereo camera. The dual stage of EKF allows the VIO solution achieve higher precision and robustness. To show the effects, we provide detailed comparisons between our DS-VIOS with other state-of-art open-source VIO approaches including OKVIS, ROVIO, VINS-MONO and S-MSCKF on the EuRoC dataset through experiments. The results demonstrate that our DS-VIO is able to achieve similar or even better performance than these state-of-art VIO approaches in term of precision and robustness.

\par

The rest of the paper is orgnized as follows: Section~\ref{Sec:Related_Work} introduces the related work. Section~\ref{Sec:First_Stage} and Section ~\ref{Sec:Second_Stage} provide the mathematical details of the first and second stage of EKF filter. Section~\ref{Sec:Dual} describes the enforcement mechanism of dual stage EKF filter. The experiments comparing our DS-VIO with state of the art open-source VIO approaches is conducted in Section~\ref{Sec:Experiments}. Finally, Section~\ref{Sec:Conclusion} draws some conclusions of this paper.

\section{RELATED WORK}
\label{Sec:Related_Work}
There are a large number of work which fuses measurements from cameras and IMUs to perform pose estimations, including VIO and VI-SLAM (Simultaneous Localization and Mapping).
The related work is discussed from three different aspects: ({\romannumeral1}) loosely or tightly coupled solutions. ({\romannumeral2}) filter-based or optimization-based solutions. ({\romannumeral3}) direct or indirect solutions.

\par

Loosely coupled solutions process the measurements from IMUs and cameras separately. Some methods process the images for computing relative-motion estimates between consecutive poses firstly \cite{Ma2012Robust,Tardif2010A,Weiss2011Real} and then fuse the result with IMU measurements. In contrast, in \cite{Matthies2012Fully,Oskiper2007Visual}, IMU measurements are used to compute rotation estimates and fuse the result with an visual estimation algorithm. Loosely coupled solutions lead to a reduction in computational cost and information loss \cite{Li2014Visual}. However, the tightly coupled solutions can achieve higher accuracy \cite{Konolige2008FrameSLAM,Konolige2010Large,mourikis2007multi,qin2018vins,sun2018robust}. In this work, we are interested in tightly coupled solutions and the proposed DS-EKF is also a tightly coupled solution.

\par

Existing tightly coupled solutions can be divided into filter-based or optimization-based solutions. The latter generally attains higher accuracy because re-linearization at each iteration can better deal with their nonlinear measurement models, like \cite{Leutenegger2014Keyframe,Lupton2012Visual,qin2018vins,Usenko2016Direct}. However, the procedure for multiple iterations lead to heavy calculation burden. As for filter-based solutions, to date, the majority of the proposed algorithms are EKF-based methods like \cite{Jones2011Visual,Kelly2008Combined,Kelly2009Visual,Kleinert2010Inertial,mourikis2007multi,Pinies2007Inertial,sun2018robust}, also, Uncented Kalman Filter in \cite{Kelly2011Visual}, and Particle Filter in \cite{yap2011particle}. In order to improve the accuracy of filter-based solutions, some researchers focus on the consistency of  estimator. The works in \cite{Huang2009A,Huang2010Observability} present the First Estimate Jacobian EKF (FEJ-EKF). Observability Constraint EKF (OC-EKF) is presented in \cite{Hesch2017Consistency,Roumeliotis2016Observability,sun2018robust}. The key idea of FEJ-EKF is to choose the same or first-ever available estimations for all the state variables as the linearization points. However, OC-EKF guarantees observability of the linearized system by ensuring the rank of nullspace of the nonlinearized system not changed after linearized. The OC-EKF is applied in our DS-VIO algorithm.

\par

The vast majority of pose estimation methods have relied on the usage of point features, which are detected and tracked in the images like \cite{mourikis2007multi,qin2018vins}. The methods in \cite{Tarrio2016Realtime,Yu2017Edge} employ edge information while the method in \cite{zheng2018pi} employs both point and line features. All the above methods are classified as indirect solutions. The methods in \cite{Forster2015IMU,Zheng2017Photometric} present a direct method which employ the image-intensity measurements directly. The direct methods exploit more information from the images than that of indirect methods.

\section{FIRST STAGE EKF FILTER DESCRIPTION}
\label{Sec:First_Stage}
The mathematical description in this paper follows the formulations in \cite{mourikis2007multi}. The gyroscope state is defined as:
\begin{eqnarray}
\textbf X_g &=&  [  {_G^I}{\overline {\textbf q} }{^T}    \quad   \textbf b{_g^T} ]^T \nonumber
\end{eqnarray}
where ${_G^I}{\overline {\textbf q} }$ is the unit quaternion describing the rotation from inertial frame (${G}$) to the body frame (${I}$), $ \textbf b{_g}\in \bbR^{3\times 1}$ describes the bias of gyroscope measurement. Here, the body frame (${I}$) is assumed to be fixed to IMU frame.

\subsection{Process Model}
\label{Sec:Model}
The state equation in the time-continuous form is given as follows:
\begin{subequations}\label{Gyro_cont_model}
\begin{eqnarray}
{_G^I}{ {\dot {  \overline {\textbf q } }}} &=& \frac{1}{2} \bm{\Omega} ( { \bm{\omega}} (t)) {_G^I}{\overline {\textbf q}}(t) \\
{\dot {\textbf  b} } _g(t) &= &{\textbf n}_{wg}(t)
\end{eqnarray}
\end{subequations}
where
$ \bm{\omega} = [\omega_x \ \omega_y \ \omega_z ]^T$ is the rotational velocity in the IMU frame,
\begin{eqnarray}
 \bm{\Omega} ( { \bm{\omega}}) =
\left[\begin{array}{cccc}
    -(\bm{\omega} \times) &    \bm{\omega}   \\
    - \bm{\omega} ^T &   0
\end{array}\right]
,
 (\bm{\omega} \times) =
\left[\begin{array}{cccc}
    0 & - \omega_z & \omega_y   \\
    \omega_z &   0 & -\omega_x  \\
   \omega_y &  \omega_x  & 0
\end{array}\right] \nonumber
\end{eqnarray}
and the gyroscope bias ${\textbf  b} _g$ is modeled as a random walk process with random walk rate $ {\textbf n}_{wg} $.
The gyroscope measurements $ \omega_m$ is:
\begin{eqnarray}
\bm{ \omega}_m =  \bm{\omega} +\textbf b_g +\textbf n_g
 \end{eqnarray}
By applying the expectation operator in \eqref{Gyro_cont_model}, we can get the estimates of the evolving gyroscope state:
\begin{subequations} \label{State_estimation}
\begin{eqnarray}
{_G^I}{\dot{ \hat{\overline {\textbf q } }  } } &=& \frac{1}{2} \bm{\Omega} ( \hat {  \bm{\omega} }  ) {_G^I}  {\hat{  \bar { \textbf  q}    }} \label{State_estimation_a} \\
{\dot{\hat {\textbf b } }} _g &=& \textbf 0_{3\times1} \label{State_estimation_b}
\end{eqnarray}
\end{subequations}
where $\hat { \bm{ \omega }}={  \bm{\omega}}_m - \hat {\textbf b}_g $.

\par
To propagate the uncertainty of the state, the discrete time state transition matrix should be computed, as for \eqref{State_estimation_a},
\begin{eqnarray}
 \nonumber 
{\Phi}_{k_q}& =&\mathrm{exp}  \biggl( \int_{t_k}^{t_{k+1}}   \frac{1}{2} \bm{\Omega} (  { \hat {\bm{\omega}}(\tau)}  ) d\tau  \biggr)_{4\times4} \approx  (\textbf I + \frac{t_k}{2} \bm{\Omega} )
\end{eqnarray}
where $t_k\define t(k)$ and $h:=t_{k+1}-t_{k}$ is the time interval between two IMU measurements. As for \eqref{State_estimation_b}, in a similar manner, one has
\begin{equation}
 \nonumber 
{\Phi}_{k_b} = \textbf 0_{3\times3}.
\end{equation}
Therefore, we can get
\begin{eqnarray}
 \nonumber 
 \textbf X_g(k+1) &=&
\left[\begin{array}{cccc}
    {_G^I}{\overline {\textbf {q}}  (k+1)}    \\
  \textbf b{_g}(k+1)
\end{array}\right]
=
\left[\begin{array}{cccc}
    {\Phi}_{k_q} {_G^I}{\overline {\textbf {q}}  (k)}   \\
   {\Phi}_{k_b} \textbf b{_g}(k)
\end{array}\right]+
\textbf w(k).
\end{eqnarray}
Therefore, one can obtain the following expression:
\begin{equation} \label{Gyro_disc_state}
\textbf X_g(k+1) \approx \textbf H_p \textbf X_g +\textbf w(k)
\end{equation}
where $\textbf H_p$ is the Jacobian of process model with respect to the gyroscope state, which is shown in Appendix, $\textbf w(k)$ is a $7 \times 1$ vector describe the process noise, of which the covariance matrix is as follows:
\[  \textbf Q_{1,k} =
\left[\begin{array}{cccc}
    \sigma{_{q}^2}\textbf I_4 & \textbf 0_{3\times 3}    \\
  \textbf 0_{4\times 4} &  \sigma{_{b}^2}\textbf I_3
\end{array}\right].
\]
Finally, the propagated covariance of the gyroscope state is described as:
\[
\textbf P_{G|G_{k+1|k}} = \textbf H_p \textbf P_{G|G_{k|k}} \textbf H_p ^T + \textbf Q_{1,k}
\]

\subsection{Measurement Model}

The first correction stage uses data from the accelerometers to correct the gyroscope state, the measurement is,
\begin{eqnarray}
\textbf z(k) &=& \textbf a(k) =  [a_x(k) \ a_y(k) \ a_z(k) ]^T
\end{eqnarray}
According to \cite{Sabatelli2013A}, $\hat {\textbf z}(k)$ can be presented by
\[
\hat{\textbf z}(k) = R_n^b
\left[\begin{array}{cccc}
   0    \\
  0    \\
  -g
\end{array}\right] =-g
\left[\begin{array}{cccc}
   2q_1q_3 - 2q_0q_2   \\
  2q_0q_1 + 2q_2q_3   \\
  q_0^2 - q_1^2 -q_2^2 +q_3^2
\end{array}\right]
\]
where $g$ is the constant g-force acceleration, and
\begin{multline}
R_n^b=
\left[ \begin{matrix}
   q_0^2 + q_1^2 -q_2^2 - q_3^2  & 2(q_1q_2+q_0q_3)   \\
   2(q_1q_2-q_0q_3)  &q_0^2 - q_1^2 + q_2^2 - q_3^2 \\
  2(q_1q_3 +q_0q_2) & 2(q_2q_3-q_0q_1)
\end{matrix} \right.
\\
\left.
\begin{matrix}
   & & 2(q_1q_3 - q_0q_2) \\
   & & 2(q_2q_3 + q_0q_1) \\
   & &q_0^2 - q_1^2 - q_2^2 + q_3^2
\end{matrix}
\right]. \nonumber
\end{multline}
The residual of the measurement can be approximated as
\begin{equation} \label{residual}
\textbf r(k) = \textbf z(k) - \hat{\textbf z}(k)  = \textbf H_g\textbf X_g +\textbf v(k)
\end{equation}
where $\textbf H_g$ is the Jacobian of measurement with respect to the gyroscope state, which is shown in Appendix, $\textbf v(k)\in \bbR^{3 \times 1}$ describes the measurement noise, of which the covariance matrix is $\sigma{_{a}^2}\textbf I_3$,

\section{SECOND STAGE EKF FILTER DESCRIPTION}
\label{Sec:Second_Stage}
The evolving IMU state is defined as follows:
\begin{eqnarray}
\nonumber 
\textbf X_I = [  {_G^I}{\overline {\textbf q} }{^T}    \quad   \textbf b{_g^T} \quad  {^G}{\textbf v}{_I^T}  \quad   \textbf b{_a^T} \quad {^G}{\textbf p}{_I^T}              ]^T
\end{eqnarray}
where $ {^G}{\textbf v}{_I} $ and $ {^G}{\textbf p}{_I}\in \mathbb{R}^{3\times 1}$ are vectors describing the IMU position and velocity in frame  ${G}$, and $  \textbf b{_a}\in \mathbb{R}^{3\times 1} $ describes the bias of accelerometer measurement. The bias is modeled as random walk processes and the random walk rate is $ {\textbf n}_{wa} $. The IMU error-state is defined as:
\begin{eqnarray}
 \widetilde{ \textbf X}_I=[  \bm{ \delta} \bm {\theta}_I^T    \quad    \widetilde{ \textbf b}_g^T     \quad     {^G}\widetilde {\textbf  v}_I^T    \quad    \widetilde {\textbf b}_a^T   \quad   {^G} \widetilde{ \textbf  p}_I^T ]^T \nonumber 
 \end{eqnarray}
where the standard additive error defined as $\tilde {x}\define x-\hat{x}$ with $ x\in\{\textbf b{_g^T},  {^G}{\textbf v}{_I^T},  \textbf b{_a^T},{^G}{\textbf p}{_I^T} \}$ and
$ \hat x\in\{\hat{ \textbf  b}{_g^T},  {^G}{\hat{ \textbf v}}{_I^T},   \hat {\textbf  b}{_a^T},{^G}{\hat {\textbf p}}{_I^T} \}$
 is used for position, velocity and bias, respectively, and $ \bm{ \delta} \bm {\theta}_I^T $ represents the quaternion error \cite{mourikis2007multi,Bloesch2015Robust}.
Ultimately, we add the $N$ camera poses in the IMU state, we can get the EKF state vector, at time-step, it can be defined as:
\[    \textbf X_E=  [ \textbf X_{I_k}^T  \quad {_G^{C_1}}{\overline {\textbf q}}^T  \quad ^G{\textbf p}{_{C_1}^T} \quad  \cdots \quad   {_G^{C_N}}{\overline {\textbf q}}^T  \quad ^G{\textbf p}{_{C_N}^T}]^T\]
where the ${_G^{C_i}}{\overline {\textbf q}}^T, {^G{\textbf p}}{_{C_i}^T}$ ,$i=1,2\cdots N$ is the camera attitude and position, respectively. The EKF error state vevtor is defined as:
\begin{eqnarray}
\tilde{\textbf X}_E&=& [ \tilde{\textbf X} _{I_k}^T \quad { \bm{\delta \theta} } {_{C_1}^T}  \quad   {^G{\widetilde{ \textbf p}}{_{C_1}^T}} \quad  \cdots \quad  { \bm{\delta \theta} } {_{C_N}^T}  \quad   {^G{\widetilde{ \textbf p}}{_{C_N}^T} }]^T \nonumber
\end{eqnarray}

\subsection{Process Model}
The continuous-time system model is described as :
\begin{subequations}\label{IMU_cont_model}
\begin{eqnarray}
{_G^I}{ {\dot {  \overline {\textbf q } }}} &=& \frac{1}{2} \Omega ( { \omega} (t)) {_G^I}{\overline {\textbf q}}(t) \\
{\dot {\textbf {  b}} } _g(t)& =& {\textbf n}_{wg}(t) \\
{^G}{\dot {\textbf  v} }_I(t)& =& {^G}{\textbf a}(t) \\
{\dot{\textbf  b} }_a(t) &= &{\textbf n}_{wa}(t) \\
{^G}\dot{\textbf{ p}} _I(t) &=& {^G}{\textbf v}_I(t)
\end{eqnarray}
\end{subequations}
where $^G{\textbf a}$ is the body acceleration in the global frame ($G$). The accelerometer measurement $ \textbf a_m$ is:
\begin{eqnarray}
\nonumber
  \textbf a_m = C({_G^I}\overline {\textbf q})(^G{\textbf a}-{^G{\textbf g}})+\textbf b_a+\textbf n_a
\end{eqnarray}
where $C(\cdot)$ denotes a rotational matrix, $\textbf n_g$ and $\textbf n_a$ are measurement noise, $^G{\textbf g}$ is gravitational acceleration expressed int the local frame.
From \eqref{Gyro_cont_model} to \eqref{Gyro_disc_state}, we can get the equations for propagating the estimates of the evolving IMU state by applying expectation operator:
\begin{subequations}\label{IMU_cont_model}
\begin{eqnarray}
{_G^I}{\dot{ \hat{\overline {\textbf q } }  } }& =& \frac{1}{2} \Omega ( \hat {  \omega}  ) {_G^I}  {\hat{  \bar { \textbf  q}    }}\\
{\dot{\hat {\textbf b } }} _g & =& \textbf 0_{3\times1} \\
{^G}{\dot {\hat{ \textbf v}}}_I &=& C({_G^I}\hat{\bar{\textbf  q } })^T {\hat {\textbf a} } +^G{\textbf g} \\
{\dot{ \hat{ \textbf b}}}_a &= &\textbf  0_{3\times1} \\
{^G}{\dot {\hat{\textbf p}}}_I &=& {^G}{\hat {\textbf v}_I}
\end{eqnarray}
\end{subequations}
where $\hat { \textbf a} =
\textbf a_m - \hat {\textbf b}_a$. Therefore, the linearized continuous-time model for the IMU error-state is:
\begin{equation} \label{linearized}
{ \dot { \widetilde {\textbf X}}_{I}  } = {\textbf F} { \widetilde {\textbf X}_{I} } +{\textbf G}{\textbf n}_{I}
\end{equation}
where ${\textbf n}_I = [ {\textbf n}{_g^T} \quad      {\textbf  n}{_{wg}^T}    \quad  {\textbf  n}{_a^T}  \quad  {\textbf n}{_{wa}^T}             ]^T$, $\textbf F$ and $\textbf G$ are shown in appendix. The state equation model can be computed as follows:
\begin{equation}
\widetilde{ \textbf X}_{E}(k+1) =  {\Phi}_k    \widetilde {\textbf X}_{E}(k) + \textbf Q_{2,k}
\end{equation}
where the computation of  transform matrix $ {\Phi}_k $ and the discrete time noise covariance $ \textbf Q_k$ can be found in \cite{sun2018robust}.
So the propagated covariance of the IMU state is described as:
\begin{eqnarray}
\nonumber
 \textbf  P_{I|I_{k+1|k}}  =  \Phi _{k}\textbf P_{I|I_{k|k}}  \Phi {_k^T}  +\textbf Q_{2,k}
\end{eqnarray}
where the covariance matrix $\textbf P_{I|I_{k|k}}$ and $\textbf  P_{I|I_{k+1|k}}$ are described as:
\begin{eqnarray}
\nonumber
   \textbf P_{E|E_{k|k}} &=&
\left[\begin{array}{cccc}
\textbf P_{I|I_{k|k}}   & \textbf   P_{I|C_{k|k}}   \\
\textbf   P_{C|I_{k|k}}  &\textbf P_{C|C_{k|k}}
\end{array}\right]\\
  \textbf  P_{E|E_{k+1|k}}& =&
\left[\begin{array}{cccc}
\textbf  P_{I|I_{k+1|k}}   &     \Phi _{k}\textbf P_{I|C_{k|k}}   \\
\textbf   P_{C|I_{k|k}}\Phi {_k^T}  &\textbf P_{C|C_{k|k}}
\end{array}\right]. \nonumber
\end{eqnarray}
When recording a new camera, we should add the camera pose into the state by using the following expressions:
\begin{eqnarray}
 _G^C{\hat {\overline{\textbf  q}}} &=& _I^C{\bar {\textbf q}} \bigotimes {_G^I}{\overline {\hat{\textbf  q}} } \nonumber \\
 ^G{\hat{\textbf  p}}_C & =& ^G{\hat{\textbf p}}_I + C({_G^I}{\hat {\bar {\textbf q}}})^T{^I}{\textbf p}_C  \nonumber
\end{eqnarray}
The augmented covariance is given by:
\begin{equation} \label{augmented_cov}
\textbf P_{k|k}  \leftarrow
\left[\begin{array}{cccc}
\textbf  I_{6N+15} \\ \textbf J
\end{array}\right]
\textbf P_{k|k}
\left[\begin{array}{cccc}
\textbf I_{6N+15} \\ \textbf J
\end{array}\right] ^T
\end{equation}
where $\textbf J$ is shown in Appendix.

\subsection{Measurement Model}
\label{Subsec:Model}
The second correction stage employs data from the images to correct the IMU state, of which the measurement model we adopted here follows \cite{mourikis2007multi,sun2018robust}. It is based the fact that static features in the wold can be observed from multiple camera poses showing constraints among all these camera poses. Consider the case of a single feature $f_j$ observed by the stereo cameras at time step $i$. Assume this feature are observed by both left and right cameras, and the left and right cameras poses are represented as $\left(_G^{C_{i,1}}\textbf q, ^G \textbf p_{C_{i,1}}\right)$ and $\left(_G^{C_{i,2}}\textbf q, ^G \textbf p_{C_{i,2}}\right)$. For the feature $f_j$ at  time step $i$, the stereo measurement is,
\begin{eqnarray}
\nonumber
 z_i^j  &=&
\left[\begin{array}{cccc}
u_{i,1}^j \\ v_{i,1}^j  \\u_{i,2}^j \\ v_{i,2}^j
\end{array}\right] =
\left[\begin{array}{cccc}
^{C_{i,1}}Z_j & \textbf 0_{2\times 2}  \\ \textbf 0_{2\times 2} &^{C_{i,2}}Z_j
\end{array}\right]
\left[\begin{array}{cccc}
^{C_{i,1}}X_j \\ ^{C_{i,1}}Y_j  \\ ^{C_{i,2}}X_j \\ ^{C_{i,2}}Y_j
\end{array}\right]
\end{eqnarray}
where $ \left[\begin{array}{cccc}  ^{C_{i,k}}X_j & ^{C_{i,k}}Y_j & ^{C_{i,k}}Y_j \end{array}\right]^T $, $k = 1,2$ represent the positions of feature $f_j$ in the left and right camera frame are given by,
\[   ^{C_i} {\textbf p}{_f{_j}}  =  \left[\begin{array}{cccc}   ^{C_{i,k}}X_j \\ ^{C_{i,k}}Y_j \\ ^{C_{i,k}}Y_j       \end{array}\right]  = C(_G^{C_{i,k}}\textbf q ) ( ^G {\textbf p} _{f_{j}}  -  ^G{\textbf  p} _{C_{i,k}}   ) \]
where $^G {\textbf p} _{f_{j}}$ is the position of feature $f_j$ in the golbal frame, it can be computed the least square method shown in \cite{mourikis2007multi}.

\par

Once $^G {\textbf p} _{f_{j}}$ is obtained, the measurement residual can be computed by:
\begin{equation}
\nonumber
{\textbf r}{_i^j}   ={\textbf  z}{_i^j} - \hat {\textbf z}{_i^j}.
\end{equation}
By linearizing about the estimates for the camera poses and feature position, the residual can be approximated as:
\[ {\textbf r}{_i^j} \approx \textbf H_{X_{i}}^j  \widetilde{ {\textbf X} }   +\textbf  H_{f_{i}}^j {^G {\widetilde{ \textbf p}} _{ f_{j}}}    + \textbf n_i^j   \]
where $\textbf n_i^j$ is the noise of the measurement, $\textbf H_{X_{i}}^j$ and $\textbf  H_{f_{i}}^j$ are the Jacobian matrixes of the measurement ${\textbf  z}{_i^j}$ with respect to the state and the feature position, respectively, of which the value of these two Jacobian matrixes can be found in \cite{sun2018robust}, ${^G {\widetilde{ \textbf p}} _{ f_{j}}}$ is the error in the position estimate of $f_j$. By stacking all the observations of feature $f_j$, one can obtain:
\begin{equation}
\nonumber
{\textbf r}{^j} \approx \textbf H_{X}^j  \widetilde{ {\textbf X}}    +\textbf  H_{f}^j {^G {\widetilde {\textbf p}} _{ f_{j}}}    + \textbf n^j.
\end{equation}
Because  the position of feature $^G {\textbf p} _{f_{j}}$ is computed by using the state estimate $\textbf X$, the error ${^G {\widetilde {\textbf p}} _{ f_{j}}}$ is correlated with the error $\widetilde {\textbf X}$. The form of residual cannot be directly used for update in the EKF. By projecting  ${\textbf r}{^j}$ to the nullspace of $\textbf  H_{f}^j$, one has
\begin{equation}
\nonumber
{\textbf r_o}{^j} = \textbf A^T{\textbf r}{^j} \approx \textbf A^T \textbf H_{X}^j  \widetilde{ {\textbf X}} + \textbf A^T \textbf n^j  = \textbf H_{X,o}^j  \widetilde {\textbf X} + \textbf n_o^j.
\end{equation}
where $\textbf A$ denotes the unitary matrix whose columns form the basis of the left nullspace of $\textbf  H_{f}^j$ \cite{mourikis2007multi}.

\section{DUAL STAGE EKF MECHANISM}
\label{Sec:Dual}
\begin{figure}[thpb]
      \centering
      \framebox{\parbox{3.3in}{\includegraphics[scale=0.36]{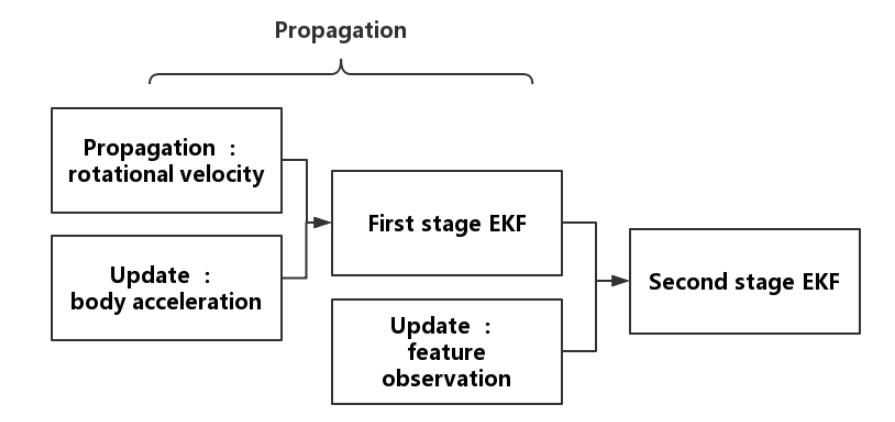}}}
      \caption{The block diagram of illustrating the structure of proposed dual stage EKF filter}
      \label{fig1_structure_block}
   \end{figure}

The structure of the proposed dual stage of EKF filter is shown in Fig.\ref{fig1_structure_block}. The first stage of EKF is designed to combine the measurements from accelerometer and gyroscope. By implementing the first stage EKF filter, data from accelerometer is used as a corrective measure by taking into
account the gravitational force to curb the error of the orientation estimate. The role of the second stage EKF is fusing measurements from IMU and stereo cameras. The images from stereo cameras can present a natural complement to IMU to to curb the errors and drift.

\par

As shown in Fig.\ref{fig1_structure_block}, the first stage of EKF is executed immediately as the IMU measurements are acquired. For each IMU measurement received, one has:
\begin{itemize}
\item \textbf {Propagation:} The rotational velocity ${\bm{\omega}_m}$ is used to propagate the gyroscope state $ \textbf X_g$, and covariance matrix $\textbf P_{G|G}$.
\item \textbf {update:} The body acceleration $\textbf a_m$ is used to perform an EKF update.
\end{itemize}

\par

The second stage of EKF contains the first stage, and the first stage is regarded as the propagate part in the second stage EKF:
\begin{itemize}
\item \textbf {Propagation:} Whenever a new IMU measurement is received, it propagates the IMU state $\textbf X_I$ and covariance matrix $\textbf P_{E|E}$.
\item \textbf {Image registration:} Whenever new stereo images are acquired, it (\romannumeral1) augments the IMU state and the corresponding covariance matrix; (\romannumeral2) operates image frontend processes, including feature extraction and feature matching.
\item \textbf {update:} EKF update is performed when (\romannumeral1) a feature has been tracked in a number of images (the number is 3 in our algorithm) is no longer detected; (\romannumeral2) camera poses will be discarded when the largest number of camera poses in the IMU state has been reached.
\end{itemize}

We present a new strategy to discard old camera poses in the above \textbf {update}. The oldest non-keyframe is discarded according to two criteria whenever new stereo images are received. The two criteria from \cite{qin2018vins} is employed here for keyframe selection. The first one is the average parallax apart from the previous keyframe and the second one is tracking quality.

\section{EXPERIMENTS}
\label{Sec:Experiments}
In this section, we perform two experiments to evaluate the proposed DS-VIO algorithm. In the first experiment, we compare the proposed algorithm with other state-of-the-art algorithms on public dataset by analyzing the Root Mean Square Error (RMSE) metrics. In the second experiment, we compare the proposed DS-VIO with S-MSCKF in the indoor environment. S-MSCKF is selected because it is also one of stereo and EKF filter based approaches.

\subsection{Dataset Comparison}

The proposed DS-VIO algorithm (stereo-filter) is compared with the state of the art approaches including OKVIS \cite{Leutenegger2014Keyframe} (stereo optimization), ROVIO \cite{Bloesch2015Robust} (monocular filter), VINS-MONO \cite{qin2018vins} (monocular optimization) and S-MSCKF \cite{sun2018robust} (stereo filter) on the EuRoC dataset \cite{Burri2016The}. These methods are different combinations of monocular, stereo, filter-based, and optimization-based methods. Among these methods, S-MSCKF is a tightly-coupled filtering-based stereo VIO algorithm which is closely related to our work. During the experiment, only the images from the left camera are used for monocular camera based algorithms like ROVIO and VINS-MONO. In order to perform a convictive experiment, only the performance of VIO is conducted and the functionality of loop closure is disabled for VINS-MONO. For each algorithm, its performance is evaluated for repeating experiments ten times and the mean value is treated as the result.

\par

The EuRoC MAV Dataset is a collection of visual-inertial datasets collected on-board a Micro Aerial Vehicle (MAV). It contains synchronized 20Hz stereo images and 200Hz IMU messages, accurate motion and structure ground-truth. Some parts of the dataset exhibit very dynamic motions and different image brightness, which renders stereo matching and feature tracking more challenging.

   \begin{figure}[h]
      \centering
      \includegraphics[width=3.2in]{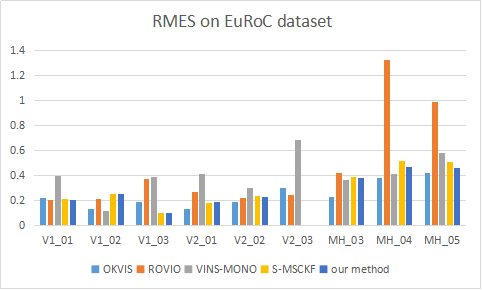}
      \caption{Root Mean Square Error (RMSE) results }
      \label{fig2_comparison}
   \end{figure}

Figure \ref{fig2_comparison} shows the RMSE results of our proposed DS-VIO algorithm and other state of the art algorithms including OKVIS, ROVIO, VINS-MONO and S-MSCKF on the EuRoC datasets. Because both our proposed method and S-MSCKF employ the KLT optical flow algorithm for feature tracking and stereo matching, they do not work properly on ``V2\_03" dataset. As pointed out in \cite{sun2018robust}, the rapid change of brightness in ``V2\_03" dataset causes failures in the stereo feature matching. On the rest datasets, our method achieves comparable performance with other methods. For the most similar method S-MSCKF, we can see that our method has better performance than that of S-MSCKF on all the datasets.

\subsection{Indoor Experiment}

\begin{figure}[b]
\centering
\subfigure[Low light]{
\begin{minipage}[t]{0.5\linewidth}
\centering
\includegraphics[width=1.5in]{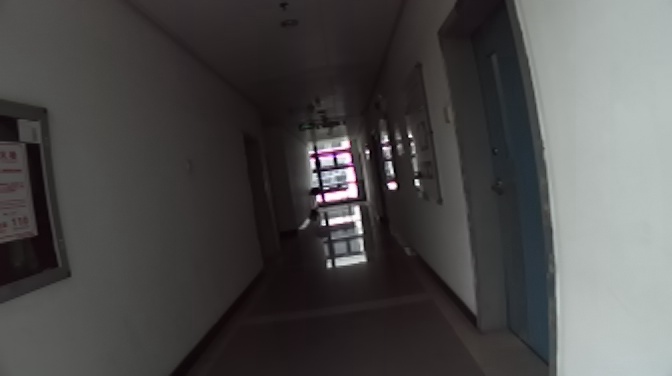}
\end{minipage}%
}%
\subfigure[Texture-less]{
\begin{minipage}[t]{0.5\linewidth}
\centering
\includegraphics[width=1.5in]{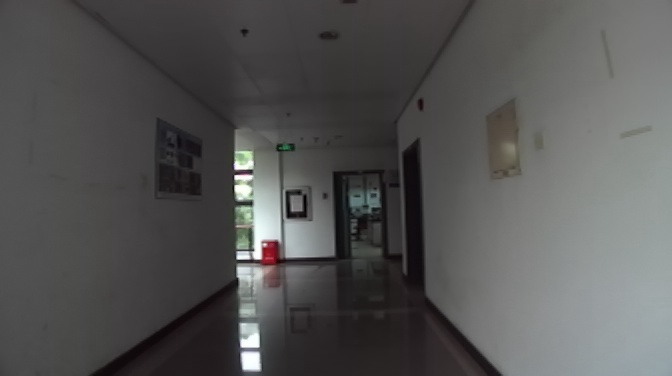}
\end{minipage}%
}%
\centering
\caption{ Images for the features of corridor environment}
\label{fig3_indoor_Environment}
\end{figure}

In the indoor experiment, for the sake of fairness, we only compare S-MSCKF with our DS-VIO because both algorithms are stereo-filter based approaches. We choose rectangular corridor in our laboratory building as the experiment area. We encounter low light and texture-less condition in the corridor environment, as shown in Fig.\ref{fig3_indoor_Environment}. The sensor suite we use is ZED-mini, which contains stereo cameras (30hz) and an IMU (800hz). With ZED-mini in our hands, we walk along the rectangular corridor for a circle around 1m/s.

\begin{figure}[htbp]
\centering

\subfigure[Trajectories in the xy plane]{
\begin{minipage}[t]{0.5\linewidth}
\centering
\includegraphics[width=1.5in]{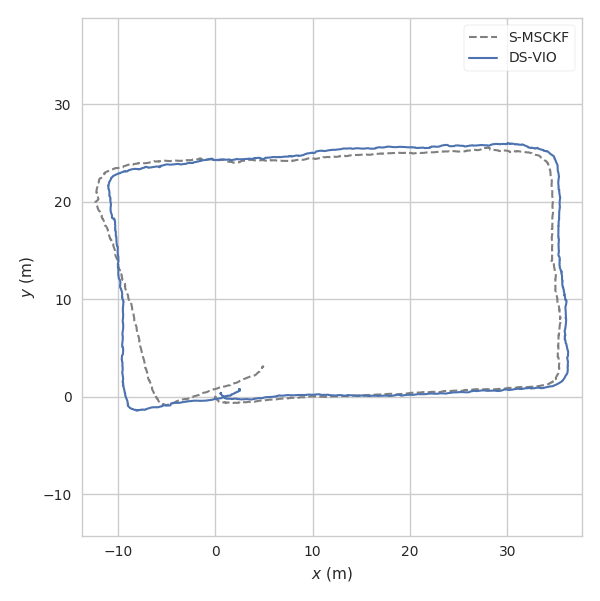}
\end{minipage}%
}%
\subfigure[Trajectories in three dimensions]{
\begin{minipage}[t]{0.5\linewidth}
\centering
\includegraphics[width=1.5in]{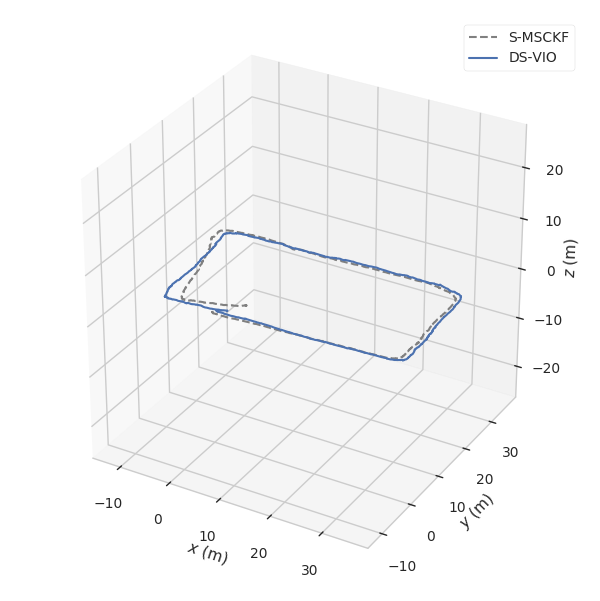}
\end{minipage}%
}%
\centering
\caption{ Results of S-MSCKF and DS-VIO}
\label{fig4_comparison_results}
\end{figure}

Fig.\ref{fig4_comparison_results} shows the comparison results between our DS-VIO and S-MSCKF. Fig.4(a) shows the trajectories in the $xy$ plane and Fig.4(b) shows the trajectories in three dimensions. The blue line represents trajectory of DS-VIO while the dotted line represents the trajectory of S-MSCKF. One can that the blue trajectory is a nonstandard rectangle but the dotted trajectory is far away from a rectangle. This comparison shows that the proposed DS-VIO achieves a smaller cumulative error than that S-MSCKF. One can attribute the archived senior performance to the first stage of EKF filter of the proposed DS-VIO, which employs the complementary characteristics between accelerometer and gyroscope.

\section{CONCLUSIONS}
\label{Sec:Conclusion}
In this paper, we present a robust and efficient filter-based stereo VIO. It employs a dual stage of EKF to perform the state estimation. This dual stage EKF filter employ the complementary characters of IMU and stereo cameras as well as accelerometer and gyroscope. The accuracy and robustness of the proposed VIO is demonstrated by experiments of the EuRoC MAV Dataset and indoor environment by comparing with the state of the art VIO algorithms. The further work should explore how to achieve better accuracy and efficiency by selecting feature points which have some certain characters.




\section*{APPENDIX}

The $\textbf H_p$ in \eqref{Gyro_disc_state} is,
\begin{equation}
\nonumber
\textbf H_p  =
\left[\begin{array}{ccccccc}
\textbf H_{p1} & \textbf H_{p2}\\
\textbf 0_{3\times4} & \textbf I_{3}
\end{array}\right]
\end{equation}
where
\begin{eqnarray}
\nonumber 
\textbf H_{p1} & =&
\left[\begin{array}{ccccccc}
1  &  -0.5\hat \omega_x t  &  -0.5\hat \omega_y t &  -0.5\hat \omega_z t      \\
 0.5\hat \omega_x t  & 1  &  0.5\hat \omega_z t &  -0.5\hat \omega_y t      \\
 0.5\hat \omega_y t  & - 0.5\hat \omega_z t  &  1 &  0.5\hat \omega_x t      \\
 0.5\hat \omega_z t  &  0.5\hat \omega_y t  &  - 0.5\hat \omega_x t   &1
\end{array}\right]
\end{eqnarray}
and
\begin{eqnarray}
\nonumber 
\textbf H_{p2} & =&
\left[\begin{array}{ccccccc}
  0.5q_1 t  & 0.5q_2 t  & 0.5q_3 t    \\
  -0.5q_0 t  & 0.5q_3 t  & -0.5q_2 t    \\
  -0.5q_3 t  & -0.5q_2 t  & 0.5q_1 t    \\
 0.5q_2 t  & - 0.5q_1 t  & -0.5q_0 t
\end{array}\right],
\end{eqnarray}
$ \hat {\bm{\omega}} = [\hat\omega_x \ \hat\omega_y \ \hat \omega_z ]^T$ is the estimation of rotational velocity.

The $\textbf H_g$ in \eqref{residual} is,
\begin{equation}
\nonumber
 \textbf H_g  =
\left[\begin{array}{ccccccc}
2gq_2 & -2gq_3 & 2gq_0 & -2gq_1 & 0&0&0\\
-2gq_1 & -2gq_0 & -2gq_3 & -2gq_2 & 0&0&0\\
-2gq_0 & 2gq_1 & 2gq_2 & 2gq_3 & 0&0&0\\
\end{array}\right]
\end{equation}
where $g$ is the g-force acceleration. The $\textbf F$ and $\textbf G$ in \eqref{linearized} are as follows:
\begin{eqnarray}
 \nonumber 
  \textbf F && =  \\
&&\left[\begin{array}{ccccc}
-(\hat { \omega} \times)  &  -{\textbf I}_3  &  {\textbf 0}_{3 \times 3} &  {\textbf 0}_{3 \times 3}  &  {\textbf 0}_{3 \times 3}\\
{\textbf 0}_{3 \times 3} &  -{\textbf I}_3  &  {\textbf 0}_{3 \times 3} &  {\textbf 0}_{3 \times 3} &  {\textbf 0}_{3 \times 3}  \\
-C({_G^I}\hat {\bar {\textbf q}})^T \cdot  (\hat {\textbf a} \times)   & \textbf 0_{3\times 3}     &  \textbf0_{3\times 3}     &- C({_G^I}\bar {\textbf q})^T   &  \textbf 0_{3\times 3}  \\
{\textbf 0}_{3 \times 3} &  -{\textbf I}_3  &  {\textbf 0}_{3 \times 3} &  {\textbf 0}_{3 \times 3} &  {\textbf 0}_{3 \times 3}  \\
{\textbf 0}_{3 \times 3} &  {\textbf 0}_{3 \times 3}  &  -{\textbf I}_3 &  {\textbf 0}_{3 \times 3} &  {\textbf 0}_{3 \times 3}
\end{array}\right] \nonumber
\end{eqnarray}
and
\begin{equation}
\nonumber
\textbf G  =
\left[\begin{array}{ccccc}
  -{\textbf I}_3  &  {\textbf 0}_{3 \times 3} &  {\textbf 0}_{3 \times 3}  &  {\textbf 0}_{3 \times 3}\\
{\textbf 0}_{3 \times 3}  &  {\textbf I}_3 &  {\textbf 0}_{3 \times 3} &  {\textbf 0}_{3 \times 3}    \\
\textbf 0_{3\times 3}     &\textbf  0_{3\times 3}     &- C({_G^I}\bar{\textbf  q})^T     & \textbf 0_{3\times 3}   \\
{\textbf 0}_{3 \times 3} &   {\textbf 0}_{3 \times 3} &  {\textbf 0}_{3 \times 3} &   {\textbf I}_3  \\
{\textbf 0}_{3 \times 3} &   {\textbf 0}_{3 \times 3} &  {\textbf 0}_{3 \times 3}  &  {\textbf 0}_{3 \times 3}
\end{array}\right].
\end{equation}
The $\textbf J$ in \eqref{augmented_cov} is follows:
\begin{equation}
\nonumber
\textbf J  =
\left[\begin{array}{ccccc}
 C({_I^C}{ {\bar {\textbf q}}})   & \textbf 0_{3\times 9}   &\textbf 0_{3\times 3}     &\textbf 0_{6N}  \\
 (C({_G^I}{ {\hat {\textbf q}}})^I{\textbf p}_I ) \times     & \textbf 0_{3\times 9}       & \textbf  I_3          & \textbf 0_{6N}
\end{array}\right].
\end{equation}

\bibliographystyle{IEEEtran}
\bibliography{VIO_Chen}

\end{document}